\documentclass[runningheads]{llncs}
\usepackage{amsmath,amssymb} 
\usepackage{color}
\usepackage{tabularx}
\usepackage{multirow}
\usepackage{cite}
\usepackage{threeparttable}
\usepackage{booktabs, caption, makecell}	
\usepackage{diagbox}
\usepackage{graphicx}

\graphicspath{{./figures/}}
\newcommand{\etal}{\textit{et al}. }
\newcommand{\ie}{\textit{i}.\textit{e}. }

\begin{document}
\title{Factorizable Net: An Efficient Subgraph-based  Framework for Scene Graph Generation} 

\titlerunning{Factorizable Net}
%
\author{Yikang LI\inst{1} \and
Wanli Ouyang\inst{2} \and
Bolei Zhou\inst{3} \and 
Jianping Shi\inst{4} \and 
Chao Zhang\inst{5} \and
Xiaogang Wang\inst{1}}
%
\authorrunning{Yikang LI \etal}
%

\institute{The Chinese University of Hong Kong, Hong Kong SAR, China \and
The University of Sydney, SenseTime Computer Vision Research Group \and 
MIT CSAIL, USA \and 
Sensetime Ltd, Beijing, China \and
Samsung Telecommunication Research Institute, Beijing, China \\
\email{\{ykli, xgwang\}@ee.cuhk.edu.hk, wanli.ouyang@sydney.edu.au, bzhou@csail.mit.edu, shijianping@sensetime.com, c0502.zhang@samsung.com}}
\maketitle              
\begin{abstract}
  Generating scene graph to describe the object interactions inside an image gains increasing interests these years. However, most of the previous methods use complicated structures with slow inference speed or rely on the external data, which limits the usage of the model in real-life scenarios. To improve the efficiency of scene graph generation, we propose a subgraph-based connection graph to concisely represent the scene graph during the inference.  A bottom-up clustering method is first used to factorize the entire graph into subgraphs, where each subgraph contains several objects and a subset of their relationships. By replacing the numerous relationship representations of the scene graph with fewer subgraph and object features, the computation in the intermediate stage is significantly reduced. In addition, spatial information is maintained by the subgraph features, which is leveraged by our proposed 
Spatial-weighted Message Passing~(SMP) structure and Spatial-sensitive Relation Inference~(SRI) module to facilitate the relationship recognition. 
On the recent Visual Relationship Detection and Visual Genome datasets, our method outperforms the state-of-the-art method in both accuracy and speed. Code has been made publicly available\footnote{https://github.com/yikang-li/FactorizableNet}.

\keywords{Visual Relationship Detection \and Scene Graph Generation \and Scene Understanding \and Object Interactions \and Language and Vision}
\end{abstract}
\section{Introduction}

\begin{figure}[t]
\includegraphics[width=\textwidth]{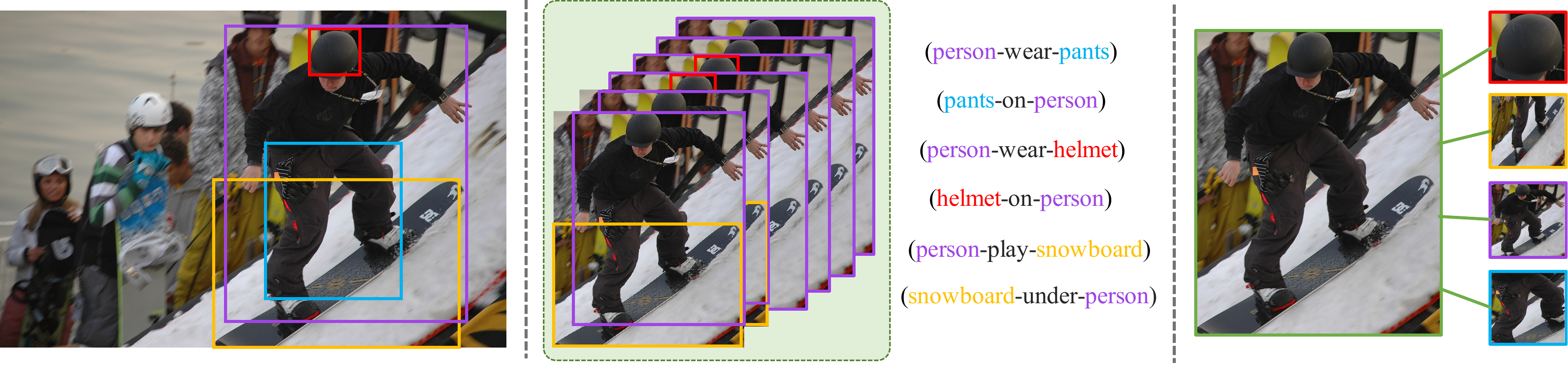}	
\caption{Left: Selected objects ; Middle: Relationships~(\emph{subject-predicate-object} triplets) are represented by phrase features in previous works~\cite{visual_relationship,visual_genome,li2017vip,li2017scene,xu2017scene,dai2017detecting,zhuang2017towards}; Right: replacing the phrases with a concise subgraph representation, where relationships can be restored with subgraph features~(green) and corresponding subject and object. }
\label{fig:motivation}
\end{figure}

Inferring the relations of the objects in images has drawn recent attentions in computer vision community on top of accurate object detection~\cite{visual_relationship,visual_genome,li2017vip,li2017scene,xu2017scene,dai2017detecting,zhuang2017towards}. Scene graph, as an abstraction of the objects and their pair-wise relationships, contains higher-level knowledge for scene understanding. Because of the structured description and enlarged semantic space of scene graphs, efficient scene graph generation will contribute to the downstream applications such as image retrieval~\cite{visual_phrase_for_retrieval, johnson2015image} and visual question answering~\cite{li2018visual, lu2018co-attending}.

Currently, there are two approaches to generate scene graphs. The first approach adopts the two-stage pipeline, which detects the objects first and then recognizes their pair-wise relationships~\cite{visual_relationship,xu2017scene,dai2017detecting,yu2017visual,liao2017natural}. 
The other approach is to jointly infer the objects and their relationships~\cite{li2017vip,li2017scene,xu2017scene} based on the object region proposals.  
To generate a complete scene graph, both approaches should group the objects or object proposals into pairs and use the features of their union area~(denoted as \emph{phrase feature}), as the basic representation for predicate inference. 
Thus, the number of phrase features determines how fast the model performs. However, due to the number of combinations growing quadratically with that of objects, the problem will quickly get intractable as the number of objects grows. Employing fewer objects~\cite{xu2017scene,li2017scene} or filtering the pairs with some simple criteria~\cite{li2017vip,dai2017detecting} could be a solution. But both sacrifice (the upper bound of) the model performance. As the most time-consuming part is the manipulations on the phrase feature, finding a more concise intermediate representation of the scene graph should be the key to solve the problem.

 We observe that multiple phrase features can refer to some highly-overlapped regions, as shown by an example in  Fig.~\ref{fig:motivation}. Prior to constructing different $\langle$subject-object$\rangle$ pairs, these features are of similar representations as they correspond to the same overlapped regions. Thus, a natural idea is to construct a shared representation for the phrase features of similar regions in the early stage. Then the shared representation is refined to learn a general representation of the are by passing the message from the connected objects. In the final stage, we can extract the required information from this shared representation to predict object relations by combining with different $\langle$subject-object$\rangle$ parirs.  Based on this observation, we propose a subgraph-based scene graph generation approach, where the object pairs referring to the similar interacting regions are clustered into a subgraph and share the phrase representation~(termed as \emph{subgraph features}). In this pipeline, all the feature refining processes are done on the shared subgraph features. 
 This design significantly reduces the number of the phrase features in the intermediate stage and speed up the model both in training and inference. 
 
As different objects correspond to different parts of the shared subgraph regions, maintaining the spatial structure of the subgraph feature explicitly retains such connections and helps the subgraph features integrate more spatial information into the representations of the region. Therefore, 2-D feature maps are adopted to represent the subgraph features. And a spatial-weighted message passing~(SMP) structure is introduced to employ the spatial correspondence between the objects and the subgraph region. 
Moreover, spatial information has been shown to be valuable in predicate recognition~\cite{dai2017detecting,liao2017natural,yu2017visual}. To leverage the such information, the Spatial-sensitive Relation Inference~(SRI) module is designed. It fuses object feature pairs and subgraph features for the final relationship inference. 
Different from the previous works, which use object coordinates or the mask to extract the spatial features, our SRI could learn to extract the embedded spatial feature directly from the subgraph feature maps.

To summarize, we propose an efficient sub-graph based scene graph generation approach with following novelties: First, a bottom-up clustering method is proposed to factorize the image into subgraphs. By sharing the region representations within the subgraph, our method could significantly reduce the redundant computation and accelerate the inference speed. In addition, fewer representations allow us to use 2-D feature map to maintain the spatial information for subgraph regions. Second, a spatial weighted message passing~(SMP) structure is proposed to pass message between object feature vectors and sub-graph feature maps. Third, a Spatial-sensitive Relation Inference~(SRI) module is proposed to use the features from subject, object and subgraph representations for recognizing the relationship between objects. Experiments on Visual Relationship Detection~\cite{visual_relationship} and Visual Genome~\cite{visual_genome} show our method outperforms the state-of-the-art method with significantly faster inference speed. Code has been made publicly available to facilitate further research.

\section{Related Work}
Visual Relationship has been investigated by numerous studies in the last decade. In the early stage, most of the works targeted on using specific types of visual relations, such as spatial relations~\cite{gupta2008beyond, johnson2015image, galleguillos2008object, choi2013understanding, kulkarni2011baby, elliott2013image} and actions~(\ie interactions between objects)~\cite{yao2010grouplet, gkioxari2015contextual, regneri2013grounding, thomason2014integrating, ramanathan2015learning, rohrbach2013translating, guadarrama2013youtube2text, antol2014zero, elhoseiny2015sherlock, farhadi2010every,xiong2015recognize}.
In most of these studies, hand-crafted features were used in relationships or phrases detection and detection works and these works were mostly supposed to leveraging other tasks, such as
object recognition~\cite{galleguillos2010context, sivic2005discovering, kumar2010efficiently, choi2010exploiting, ladicky2010graph, salakhutdinov2011learning, rabinovich2007objects, fidler2007towards, russell2006using},
image classification and retrieval~\cite{mensink2014costa, gong2014multi},
scene understanding and generation~\cite{zitnick2013learning, hoiem2008putting, chang2014semantic, yao2012describing, izadinia2014incorporating, gould2008multi, berg2012understanding},
as well as text grounding\cite{plummer2015flickr30k, karpathy2014deep, rohrbach2015grounding}.
However, in this paper, we focus on the higher-performed method dedicated to \emph{generic} visual relationship detection task which is essentially different from  works in the early stage.

In recent years, new methods are developed specifically for detecting visual relationships. 
An important series of methods~\cite{das2013thousand, divvala2014learning, sadeghi2011recognition}
consider the \emph{visual phrase} as an integrated whole, \ie considering each distinct combination of object categories and relationship predicates as
a distinct class. 
Such methods will become intractable when the number of such combinations becomes very large. 

As an alternative paradigm, considering relationship predicates and object categories separately becomes more popular in recent works~\cite{liao2017natural,peyre2017weakly,zhang2017visual,zhuang2017towards}. 
Generic visual relationship detection was first introduced as a visual task by Lu \etal in \cite{visual_relationship}. In this work, objects are detected first, and then the predicates between object pairs are recognized, where word embeddings of the object categories are employed as language prior for predicate recognition. 
Dai \etal proposed DR-Net to exploit the statistical dependencies between objects and their relationships for this task~\cite{dai2017detecting}. In this work, a CRF-like optimization process is adopted to refine the posterior probabilities iteratively~\cite{dai2017detecting}. 
Yu \etal presented a Linguistic Knowledge Distillation pipeline to employ the annotations and external corpus~(\ie wikipedia), where strong correlations between predicate and $\langle$subject-object$\rangle$ pairs are learned to regularize the training and provide extra cues for inference~\cite{yu2017visual}. 
Plummer \etal designed a large collection of handcrafted linguistic and visual cues for visual relationship detection and constructed a pipeline to learn the weights for combining them~\cite{plummer2016phrase}.
Li~\etal used the message passing structure among subject, object and predicate branches to model their dependencies~\cite{li2017vip}. 

The most related works are the methods proposed by Xu~\etal~\cite{xu2017scene} and Li~\etal~\cite{li2017scene}, both of which jointly detect the objects and recognize their relationships.  
In \cite{xu2017scene}, the scene graph was constructed by refining the object and predicate features jointly in an iterative way. In \cite{li2017scene}, region caption was introduced as a higher-semantic-level task for scene graph generation, so the objects, pair-wise relationships and region captions help the model learn representations from three different semantic levels. 
Our method differs in two aspects:
(1) We propose a more concise graph to represent the connections between objects instead of enumerating every possible pair, which significantly reduces the computation complexity and allows us to use more object proposals; 
(2) Our model could learn to leverage the spatial information embedded in the subgraph feature maps to boost the relationship recognition.  
Experiments show that the proposed framework performs substantially better and faster in 
all different task settings. 

\begin{figure}[t]
	\includegraphics[width=\textwidth]{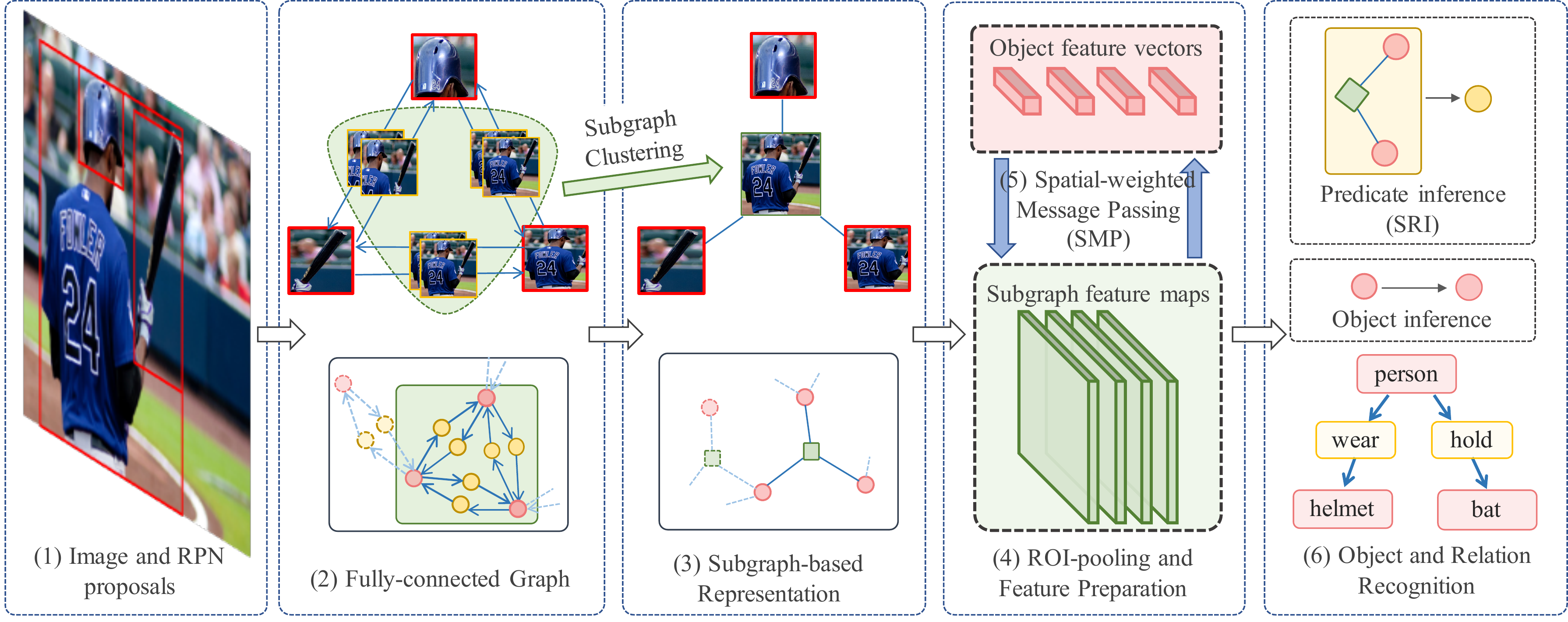}
	\caption{Overview of our F-Net. (1) RPN is used for object region proposals, which shares the base CNN with other parts. (2) Given the region proposal, objects are grouped into pairs to build up a fully-connected graph, where every two objects are connected with two directed edges. (3) Edges which refer to similar phrase regions are merged into subgraphs, and a more concise connection graph is generated. (4) ROI-Pooling is employed to obtain the corresponding features~(2-D feature maps for subgraph and feature vectors for objects). (5) Messages are passed between subgraph and object features along the factorized connection graph for feature refinement. (6) Objects are predicted from the object features and predicates are inferred based on the object features and the subgraph features. Green, red and yellow items refer to the subgraph, object and predicate respectively.}
	\label{fig:framework}
\end{figure}

\section{Framework of the Factorizable Network}

The overview of our proposed Factorizable Network~(F-Net) is shown in Figure~\ref{fig:framework}. Detailed introductions to different components will be given in the following sections.

The entire process can be summarized as the following steps: (1) generate object region proposals with Region Proposal Network~(RPN)~\cite{faster_rcnn}; (2) group the object proposals into pairs and establish the fully-connected graph, where every two objects have two directed edges to indicate their relations; (3) cluster the fully-connected graph into several subgraphs and share the subgroup features for object pairs within the subgraph, then a factorized connection graph is obtained by treating each subgraph as a node; (4) ROI pools~\cite{fast_rcnn,he2017mask} the objects and subgraph features and transforms them into feature vectors and 2-D feature maps respectively; (5) jointly refine the object and subgraph features by passing message along the subgraph-based connection graph for better representations; (6) recognize the object categories with object features and their relations~(predicates) by fusing the subgraph features and object feature pairs. 

\subsection{Object Region Proposal}
Region Proposal Network~\cite{faster_rcnn} is adopted to generate object proposals. It shares the base convolution layers with our proposed F-Net. An auxiliary convolution layer is added after the shared layers. The anchors are generated by clustering the scales and ratios of ground truth bounding boxes in the training set~\cite{li2017scene}. 

\subsection{Grouping Proposals into Fully-connected Graph}
As every two objects possibly have two relationships in opposite directions, we connect them with two directed edges~(termed as phrases). A fully-connected graph is established, where every edge corresponds to a potential relationship~(or \emph{background}). Thus, $N$ object proposals will have $N(N-1)$ candidate relations~(yellow circles in Fig.~\ref{fig:framework} (2)). Empirically, more object proposals will bring higher recall and make it more likely to detect objects within the image and generate a more complete scene graph. However, large quantities of candidate relations may deteriorate the model inference speed. Therefore, we design an effective representations of all these relationships in the intermediate stage to adopt more object proposals.

\subsection{Factorized Connection Graph Generation}\label{sec:subgraph}
By observing that many relations refer to overlapped regions~(Fig.~\ref{fig:motivation}), we share the representations of the phrase region to reduce the number of the intermediate phrase representations as well as the computation cost. 
For any candidate relation, it corresponds to the union box of two objects~(the minimum box containing the two boxes). Then we define its confidence score as the product of the scores of the two object proposals. With confidence scores and bounding box locations, non-maximum-suppression~(NMS)~\cite{fast_rcnn} can be applied to suppress the number of the similar boxes and keep the bounding box with highest score as the representative. So these merged parts compose a subgraph and share an unified representation to describe their interactions. 
Consequently, we get a subgraph-based representation of the fully-connected graph: every subgraph contains several objects; every object belongs to several subgraphs; every candidate relation refers to one subgraph and two objects. 

\noindent\textbf{Discussion} In previous work, ViP-CNN~\cite{li2017vip} proposed a triplet NMS to preprocess the relationship candidates and remove some overlapped ones. However, it may falsely discard some possible pairs because only spatial information is considered. Differently, our method just proposes a concise representation of the fully-connect graph by sharing the intermediate representation. It does not prune the edges, but represent them in a different form. Every predicate will still be predicted in the final stage. Thus, it is no harm for the model potential to generate the full graph.

\subsection{ROI-Pool the Subgraph and Object Features}
After the clustering, we have two sets of proposals: objects and subgraphs. Then ROI-pooling~\cite{fast_rcnn,he2017mask} is used to generate corresponding features. Different from the prior art methods~\cite{xu2017scene,li2017scene} which use feature vectors to represent the phrase features, we adopt 2-D feature maps to maintain the spatial information within the subgraph regions. As the subgraph feature is shared by several predicate inferences, 2-D feature map can learn more general representation of the region and its inherit spatial structure can help to identify the subject/object and their relations, especially the spatial relations. We continue employing the feature vector to represent the objects. Thus, after the pooling, 2-D convolution layers and fully-connected layers are used to transform the subgraph feature and object features respectively.

\begin{figure}[t]
	\includegraphics[width=\textwidth]{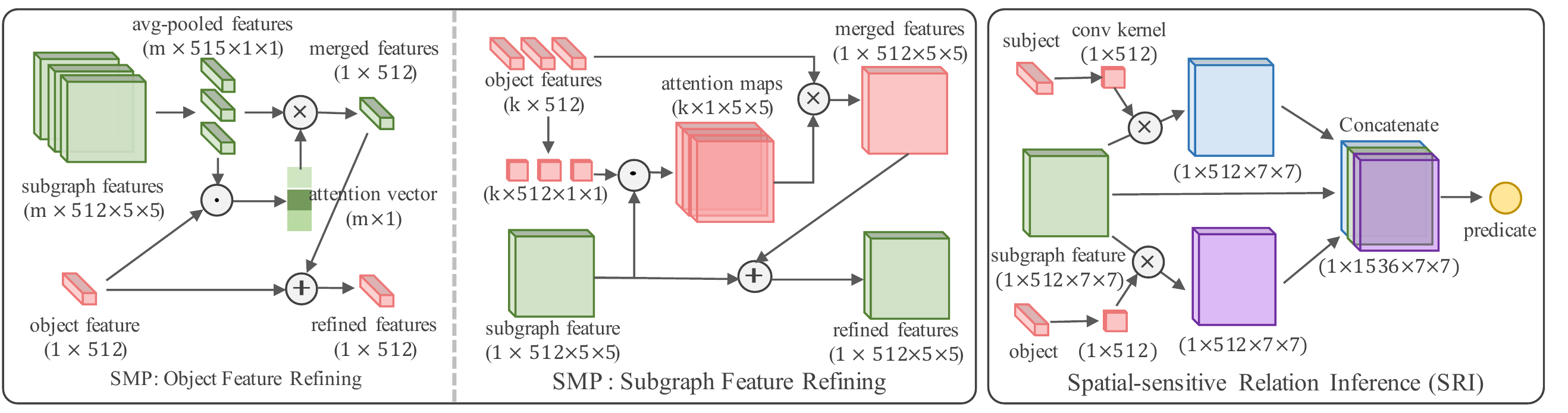}
	\caption{Left:SMP structure for object/subgraph feature refining. Right: SRI Module for predicate recognition. Green, red and yellow  refer to the subgraphs, objects and predicates respectively. $\odot$ denotes the dot product. $\oplus$ and $\otimes$ denote the element-wise sum and product.}
	\label{fig:modules}
\end{figure}

\subsection{Feature Refining with Spatial-weighted Message Passing} 
As object and subgraph features involve different semantic levels, where objects concentrate on the details and subgraph focus on their interactions, passing message between them could help to learn better representations by leveraging their complementary information. Thus, we design a spatial weighted message passing~(SMP) structure to pass message between object feature vectors and subgraph feature maps~(left part of Fig.~\ref{fig:modules}). Messages passing from objects to subgraphs and from subgraphs to objects are two parallel processes. $\mathbf{o_i}$ denotes the object feature vector and $\mathbf{S_k}$ denotes the subgraph feature map. 

\subsubsection{Pass Message From Subgraphs to Objects}

This process is to pass several 2-D feature maps to feature vectors. 
Since objects only require the general information about the subgraph regions instead of their spatial information, 2-D average pooling is directly adopted to pool the 2-D feature maps $\mathbf{S_k}$ into feature vectors $\mathbf{s_k}$.  
Because each object is connected to various number of subgraphs, we need first aggregate the subgraph features and then pass them to the target object nodes. Attention~\cite{show_attend_tell} across the subgraphs is employed to keep the scale aggregated features invariant to the number of input subgraphs and determine the importance of different subgraphs to the object:

\begin{equation}
\small 
\tilde{\mathbf{s}}_i=\sum_{\mathbf{S_k}\in \mathbb{S}_i} p_i(\mathbf{S}_k) \cdot \mathbf{s}_k
\label{eq:message_pass_s2o}
\end{equation}

\noindent where $\mathbb{S}_i$ denotes the set of subgraphs connected to object $i$. $\tilde{\mathbf{s}}_i$ denotes aggregated subgraph features passed to object $i$. $\mathbf{s}_k$ denotes the feature vector average-pooled from the 2-D feature map $\mathbf{S}_k$. $p_i(\mathbf{S}_k)$ denotes the probability that $\mathbf{s}_k$ is passed to the target $i$-th object~(attention vector in Fig.~\ref{fig:modules}):

\begin{equation}\small
\small 
p_i(\mathbf{S}_k)=\frac{\exp{\left(\mathbf{o}_i \cdot \mathrm{FC}^{(att\_s)}\left(\mathrm{ReLU}\left(\mathbf{s}_k\right)\right)\right)}}{
		\sum_{\mathbf{S_k}\in \mathbb{C}_i} 
			\exp{\left(\mathbf{o}_i \cdot \mathrm{FC}^{(att\_s)}\left(\mathrm{ReLU}\left(\mathbf{s}_k\right)\right)\right)}}
\label{eq:message_pass_s2o_p}
\end{equation}
\noindent where $\mathrm{FC}^{(att\_s)}$ transforms the feature $\mathbf{s}_k$ to the target domain of $\mathbf{o}_i$. ReLU denotes the Rectified Linear Unit layer~\cite{relu}. 

After obtaining message features, the target object feature is refined as:
\begin{equation}\small
\hat{\mathbf{o}}_i=\mathbf{o}_i +  \textrm{FC}^{(s\rightarrow o)}\left(\textrm{ReLU}\left(\tilde{\mathbf{s}}_i\right)\right)
\label{eq:message_pass_s2o_refine}
\end{equation}
\noindent where $\hat{\mathbf{o}}_i$ denotes the refined object feature. $\textrm{FC}^{\left(s\rightarrow o\right)}$ denotes the fully-connected layer to transform merged subgraph features to the target object domain.

\subsubsection{Pass Message From Objects to Subgraphs}
Each subgraph connects to several objects, so this process is to pass several feature vectors to a 2-D feature map. Since different objects correspond to different regions of the subgraph features, when aggregating the object features, their weights should also depend on their locations:

\begin{equation}\small
\tilde{\mathbf{O}}_k\left(x,y\right)=\sum_{\mathbf{o_i}\in \mathbb{O}_k} \mathbf{P}_k(\mathbf{o}_i)( x, y) \cdot \mathbf{o}_i
\label{eq:message_pass_o2s}
\end{equation}

\noindent where $\mathbb{O}_k$ denotes the set of objects contained in subgraph $k$. $\tilde{\mathbf{O}}_k\left(x,y\right)$ denotes aggregated object features to pass to subgraph $k$ at location~$\left(x,y\right)$. $\mathbf{P}_k(\mathbf{o}_i)(x,y)$ denotes the probability map that the object feature $\mathbf{o}_i$ is passed to the $k$-th subgraph at location $(x,y)$~(corresponding to the \emph{attention maps} in Fig.~\ref{fig:modules}):

\begin{equation}\small
\mathbf{P}_k(\mathbf{o}_i)(x,y)=\frac{\exp{\left(\mathrm{FC}^{(att\_o)}\left(\mathrm{ReLU}\left(\mathbf{o}_i\right)\right) \cdot \mathbf{S}_k(x,y)\right)}}{
		\sum_{\mathbf{S_k}\in \mathbb{C}_i} 
			\exp{\left(\mathrm{FC}^{(att\_o)}\left(\mathrm{ReLU}\left(\mathbf{o}_i\right)\right) \cdot \mathbf{S}_k(x,y)\right)}}
\label{eq:message_pass_o2s_p}
\end{equation}

\noindent where $\mathrm{FC}^{(att\_o)}$ transforms $\mathbf{o}_i$ to the target domain of $\mathbf{S}_k(x,y)$. 
The probabilities are summed to 1 across all the objects at each location to normalize the scale of the message features. But there are no such constraints along the spatial dimensions. So different objects help to refine different parts of the subgraph features. 

After the aggregation in Eq.~\ref{eq:message_pass_o2s}, we get a feature map where the object features are aggregated with different weights at different locations. Then we can refine the subgraph features as:

\begin{equation}\small
\hat{\mathbf{S}}_k=\mathbf{S}_k + \textrm{Conv}^{(o\rightarrow s)}\left(\textrm{ReLU}\left(\tilde{\mathbf{O}}_k\right)\right)
\label{eq:message_pass_o2s_refine}
\end{equation}

\noindent where $\hat{\mathbf{S}}_i$ denotes the refined subgraph features. $\textrm{Conv}^{\left(o\rightarrow s\right)}$ denotes the convolution layer to transform merged object messages to the target subgraph domain. 

\noindent\textbf{Discussion} Since subgraph features embed the interactions among several objects and objects are the basic elements of subgraphs, message passing between object and subgraph features could: (1) help the object feature learn better representations by considering its interactions with other objects and introduce the contextual information; (2) refine different parts of subgraph features with corresponding object features. 
Different from the message passing in ISGG~\cite{xu2017scene} and MSDN~\cite{li2017scene}, our SMP (1) passes message between ``points''~(object vectors) and ``2-D planes''~(subgraph feature maps); (2) adopts attention scheme to merge different messages in a normalized scale. 
Besides, several SMP modules can be stacked to enhance the representation ability of the model. 

\subsection{Spatial-sensitive Relation Inference}

After the message passing, we have got refined representations of the objects $\mathbf{o}_i$ and subgraph regions $\mathbf{S}_k$. Object categories can be predicted directly with the object features. Because subgraph features may refer to several object pairs, we use the subject and object features along with their corresponding subgraph feature to predict their relationship:
\begin{equation}\small
\mathbf{p}^{\langle i,k,j\rangle} = \mathbf{f}\left(\mathbf{o}_i, \mathbf{S}_k, \mathbf{o}_j\right)
\label{eq:predicate}
\end{equation}

As different objects correspond to different regions of subgraph features, 
subject and object features work as the convolution kernels to extract the visual cues of their relationship from feature map. 
\begin{equation}\small
\mathbf{S}^{(i)}_{k} = \mathrm{FC}\left(\mathrm{ReLU}\left(\mathbf{o}_i\right)\right) \otimes \mathrm{ReLU}\left(\mathbf{S}_k\right)
\label{eq:conv_s}
\end{equation}
\noindent where $\mathbf{S}^{(i)}_{k}$ denotes the convolution result of subgraph feature map $\mathbf{S}_k$ with $i$-th object as convolution kernel. $\otimes$ denotes the convolution operation. As learning a convolution kernel needs large quantities of parameters, Group Convolution~\cite{imagenet_hinton} is adopted. We set group numbers as the number of channels, so the group convolution can be reformulated as element-wise product. 

Then we concatenate $\mathbf{S}^{(i)}_{k}$ and $\mathbf{S}^{(j)}_{k}$ with the subgraph feature $\mathbf{S}_{k}$ and predict the relationship directly with a fully-connected layer:

\begin{equation}\small
\mathbf{p}^{\langle i,k,j\rangle} = \mathrm{FC}^{(p)}\left(\mathrm{ReLU}\left(\left[\mathbf{S}^{(i)}_{k};\mathbf{S}_{k};\mathbf{S}^{(j)}_{k}\right]\right)\right)
\label{eq:predicate_fc}
\end{equation}

\noindent where $\mathrm{FC}^{(p)}$ denotes the fully-connected layer for predicate recognition. $\left[\cdot\right]$ denotes the concatenation. 

\noindent\textbf{Bottleneck Layer}
Directly predicting the convolution kernel leads to a lot of parameters to learn, which makes the model huge and hard to train. The number of parameters of $\mathrm{FC}^{(p)}$ equals:
\begin{equation}\small
\#\mathrm{FC}^{(p)} = C^{(p)} \times C \times W\times H
\label{eq:para_size}
\end{equation}
\noindent where $C^{(p)}$ denotes the number of predicate categories. $C$ denotes the channel size. $W$ and $H$ denote the width and height of the feature map. 
Inspired by the bottleneck structure in ~\cite{resnet}, we introduce an additional $1\times1$ bottleneck convolution layer prior to $\mathrm{FC}^{(p)}$ to reduce the number of channels~(omitted in Fig.~\ref{fig:modules}).
After adding an bottleneck layer with channel size equalling to $C^\prime$, the parameter size gets:
\begin{equation}\small
\#\mathrm{Conv}^{(bottleneck)} + \#\mathrm{FC}^{(p)}  = C\times C^\prime + C^{(p)} \times C^\prime \times W\times H
\label{eq:para_size_bottleneck}
\end{equation}
If we take $C^\prime=C/2$, as $\#\mathrm{Conv}^{(bottleneck)}$ is far less than $\#\mathrm{FC}^{(p)}$, we almost half the number of parameters. 

\noindent\textbf{Discussion}
In previous work, spatial features have been extracted from the coordinates of the bounding box or object masks~\cite{dai2017detecting,liao2017natural,yu2017visual}. Different from these methods, ours embeds the spatial information in the subgraph feature maps. Since $\mathrm{FC}^{(p)}$ has different weights at different locations, it could learn to decide whether to leverage the spatial feature and how to use that by itself from the training data. 

\section{Experiments}
In this section, implementation details of our proposed method and experiment settings will be introduced. Ablation studies will be done to show the effectiveness of different modules. We also compare our F-Net with state-of-the-art methods on both accuracy and testing speed. 

\subsection{Implementation details}
\subsubsection{Model details} ImageNet pretrained VGG16~\cite{VGG} is adopted to initialize the base CNN, which is shared by RPN and F-Net. ROI-align~\cite{he2017mask} is used to generated $5 \times 5$ object and subgraph features. Two FC layers are used to transform the pooled object features to 512-dim feature vectors. Two $3\times 3$ Conv layers are used to generate 512-dim subgraph feature maps. For SRI module, we use a 256-dim bottleneck layer to reduce the model size. All the newly introduced layers are randomly initialized. 

\subsubsection{Training details} 
During training, we fix Conv$_1$ and Conv$_2$ of VGG16, and set the learning rate of the other convolution layers of VGG as 0.1 of the overall learning rate. Base learning rate is 0.01,  and get multiplied by 0.1 every 3 epochs. RPN NMS threshold is set as 0.7. Subgraph clustering threshold is set as 0.5. For the training samples, 256 object proposals and 1024 predicates are sampled 50\% foregrounds. There is no sampling for the subgraphs, so the subgraph connection maps are identical from training to testing. The RPN part is trained first, and then RPN, F-Net and base VGG part are jointly trained. 

\subsubsection{Inference details}
During testing phase, RPN NMS threshold and subgraph clustering threshold are set as 0.6 and 0.5 respectively. All the predicates~(edges of fully-connected graph) will be predicted. Top-1 categories will be used as the prediction for objects and relations.  Predicated relationship triplets will be sorted in the descending order based on the products of their subject, object and predicate confidence probabilities. Inspired by Li~\etal in \cite{li2017vip}, triplet NMS is adopted to remove the redundant predictions if the two triplets refer to the identical relationship. 
\begin{table*}[t]
	\renewcommand{\arraystretch}{1.1}
	\setlength{\tabcolsep}{3.8pt}
	\small
	\caption{Dataset statistics. \textbf{VG-MSDN} and \textbf{VG-DR-Net} are two cleansed-version of raw Visual Genome dataset. \textbf{\#Img} denotes the number of images. \textbf{\#Rel} denotes the number of subject-predicate-object relation pairs. \textbf{\#Object} and \textbf{\#Predicate} denotes the number of object and predicate categories respectively. }.
	\begin{center}
	\begin{tabular}{l|cc | cc | c|c}
	\hline
	\multirow{2}{*}{Dataset} & \multicolumn{2}{c}{Training Set} \vline & \multicolumn{2}{c}{Testing Set}\vline & \multirow{2}{*}{\#Object} & \multirow{2}{*}{\#Predicate	} \\
	& \#Img & \#Rel & \#Img & \#Rel & \\\hline
	VRD~\cite{visual_relationship} & 4,000 & 30,355 & 1,000 & 7,638 & 100 & 70 \\
	VG-MSDN~\cite{li2017scene,visual_genome} & 46,164 & 507,296 & 10,000 & 111,396 & 150 & 50 \\
	VG-DR-Net~\cite{dai2017detecting,visual_genome} & 67,086 & 798,906 & 8,995 & 26,499 & 399 & 24\\
	\hline
	\end{tabular}
	\end{center}
	\label{tab:dataset}
\end{table*}

\subsection{Datasets}

Two datasets are employed to evaluate our method, Visual Relationship Detection~(VRD)~\cite{visual_relationship} and Visual Genome~\cite{visual_genome}. VRD is a small benchmark dataset where most of the existing methods are evaluated. Compared to VRD, raw Visual Genome contains too many noisy labels, so dataset cleansing should be done to make it available for model training and evaluation. For fair comparison, we adopt two cleansed-version Visual Genome used in \cite{li2017scene} and \cite{dai2017detecting} and compare with their methods on corresponding datasets. Detailed statistics of the three datasets are shown in Tab.~\ref{tab:dataset}.

\subsection{Evaluation Metrics}
Models will be evaluated on two tasks, \emph{Visual Phrase Detection~(PhrDet)} and \emph{Scene Graph Generation~(SGGen)}. Visual Phrase Detection is to detect the $\langle$subject-predicate-object$\rangle$ phrases, which is tightly connected to the Dense Captioning~\cite{densecap}. Scene Graph Generation is to detect the objects within the image and recognize their pair-wise relationships. Both tasks recognize the $\langle$subject-predicate-object$\rangle$ triplets, but scene graph generation needs to localize both the subject and the object with at least 0.5 IOU~(intersection over union) while visual phrase detection only requires one bounding box for the entire phrase.  

Following~\cite{visual_relationship}, Top-K Recall~(denoted as \emph{Rec@K}) is used to evaluate how many labelled relationships are hit in the top K predictions. The reason why we use Recall instead of mean Average Precision~(mAP) is that annotations of the relationships are not complete. mAP will falsely penalize the positive but unlabeled relations. In our experiments, \emph{Rec@50} and \emph{Rec@100} will be reported.

The testing speed of the model is also reported. Previously, only accuracy is reported in the papers. So lots of complicated structure and post-processing methods are used to enhance the Recall. 
As scene graph generation is getting closer to the practical applications and products, testing speed become a critical metric to evaluate the model. If not specified, testing speed is evaluated with Titan-X GPU.

\begin{table*}[t]
	\renewcommand{\arraystretch}{1.1}
	\setlength{\tabcolsep}{3.5pt}
	\small
	\caption{Ablation studies of the proposed model. \textbf{PhrDet} denotes phrase detection task. \textbf{SGGen} denotes the scene graph generation task. \textbf{SubGraph} denotes whether to use Subgraph-based clustering strategy. \textbf{2-D} indicates whether we use 2-D feature map or feature vector to represent subgraph features. \textbf{\#SMP} denotes the number of the Multimodal Message Passing structures~(model 1 adopts the message passing in \cite{li2017scene}).  \textbf{\#Boxes} denotes the number of object proposals we use during the testing. \textbf{SRI} denotes whether the SRI module is used~(baseline method is average pooling the subgraph feature maps to vectors). \textbf{Speed} shows the time spent for one inference forward pass~(second/image).}
	\begin{center}
		\begin{tabularx}{1.0\linewidth}{c|ccccc | cccc | c}
			\hline
			\multirow{2}{*}{ID} & \multirow{2}{*}{SubGraph} & \multirow{2}{*}{\#SMP} & \multirow{2}{*}{2-D} & \multirow{2}{*}{SRI} & \multirow{2}{*}{\#Boxes} & \multicolumn{2}{c}{\textbf{PhrDet}} & \multicolumn{2}{c}{\textbf{SGGen}}\vline & \multirow{2}{*}{\textbf{Speed}}\\
			& & & & & & R@50& R@100 & R@50& R@100 &\\
			\hline
			0 &-  & 0 & - & -      		& 64 &  16.92 & 21.04 & 8.52 & 10.81  	& 0.65\\
			1 & \checkmark & 0 & - & - & 64	 &  16.50 & 20.79 & 8.49 & 10.33 	& 0.18 \\
			2 & \checkmark & 0 & - & - & 200 &  18.71 & 22.77 & 9.73 & 12.02 	& 0.20  \\
			3 & \checkmark & 0 & \checkmark 	& - & 200 &  19.09 & 22.88 & 9.90 & 12.08 & 0.32  \\
			4 & \checkmark & 1 & \checkmark 	& - & 200 &  20.48 & 25.69 & 11.62 & 14.55 & 0.42 \\
			5 & \checkmark & 1 & \checkmark 	& \checkmark & 200 & 22.54  & 28.31 & 12.83 & 16.12 & 0.44 \\
			6 & \checkmark & 2 & \checkmark & \checkmark & 200	& 22.84  & 28.57 & 13.06 & 16.47 & 0.55  \\
			\hline
		\end{tabularx}
	\end{center}
	\label{tab:component}
\end{table*}

\subsection{Component Analysis}\label{sec:ablation}
In this section, we perform several experiments to evaluate the effectiveness of different components of our F-Net~(Tab.~\ref{tab:component}). All the experiments are performed on VG-MSDN~\cite{li2017scene} as it is larger than VRD~\cite{visual_relationship} to eliminate overfitting and contains more predicate categories than VG-DR-Net~\cite{dai2017detecting}.

\subsubsection{Subgraph-based pipeline}
For the baseline model 0, every relation candidate is represented by a phrase feature vector, and the predicates are predicted based on the concatenation of subject, object and phrase features. In comparison, model 1 and 2 adopt the subgraph-based presentation of the fully-connected graph with different numbers of object proposals. 
By comparing model 0 and 1, we can see that subgraph-based clustering could significantly speed up the model inference because of the fewer intermediate features. However, since most of the phrase features are approximated by the subgraph features, the accuracy of model 1 is slightly lower than that of model 0.
However, the disadvantage of model 1 can be easily compensated by employing more object proposal as model 2 outperforms model 0 both in speed and accuracy by a large scale. Furthermore, model 1$\sim$6 are all faster than model 0, which proves the efficiency of our subgraph-based representations.

\subsubsection{2-D feature map}
From model 3, we start to use 2-D feature map to represent the subgraph features, which can maintain the spatial information within the subgraph regions. Compared to model 2, model 3 adopts 2-D representations of the subgraph features and use average-pooled subgraph features~(concatenated with the subject and object feature) to predict the relationships. Since SRI is not used, the main difference is two $3\times3$ conv layers are used instead of FC layer to transform the subgraph features. Since we pool the subgraph regions to $5\times 5$ feature maps, which is just the perceptual field of two $3\times3$ conv layers, therefore, model 3 has less parameters to learn and the spatial structure of the feature map could serve as a regularization. Therefore, compared to model 2, model 3 performs better. 

\subsubsection{Message Passing between objects and subgraphs} When comparing Model 3 and 4, 2.02\%$\sim$2.37\% SGGen Recall increase is observed, which shows our proposed SMP could also help the model learn a better representation of the objects and the subgraph regions. With our proposed SMP, different parts of subgraph features can be refined by different objects, and object features can also get refined by receiving more information about their interactions with other object regions. 
Furthermore, when comparing model 5 and model 6, we can see that stacking more SMP modules can further improve the model performance as more complicated message paths are introduced. However, more SMP modules will deteriorate the testing speed, especially when we use feature maps to represent the subgraph features.

\subsubsection{Spatial-sensitive Relation Inference}
From Eq.~\ref{eq:predicate_fc}, Fully-Connected layer is used to predict the relationships from the 2-D feature map, so every point within the map will be assigned a location-specified weight and the SRI could learn to model the hidden spatial connections. Different from previous models that employing handcrafted spatial features like axises of the subject/object proposals, our model could not only improve the recognition accuracy of explicit spatial relationships like \emph{above} and \emph{below}, but also learn to extract the inherit spatial connection of other relationships. Experiment results of model 4 and 5 show the improvement brought by our proposed SRI module. 

\begin{table*}[t]
	\renewcommand{\arraystretch}{1.1}
	\setlength{\tabcolsep}{3.pt}
	\small
	\caption{Comparison with existing methods on visual phrase detection~(\textbf{PhrDet}) and scene graph generation(\textbf{SGGen}). \textbf{Speed} indicates the testing time spent on one image~(second/image). Benchmark dataset, VRD~\cite{visual_relationship}, and two cleansed-version Visual Genome~\cite{visual_genome,li2017scene,dai2017detecting} are used for fair comparison.}
	\begin{center}
		\begin{tabularx}{1.0\linewidth}{l | l | cc | cc | c}
		\hline
		\multirow{2}{*}{Dataset} & \multirow{2}{*}{Model} & \multicolumn{2}{c}{\textbf{PhrDet}} \vline & \multicolumn{2}{c}{\textbf{SGGen}} \vline  &  \multirow{2}{*}{\textbf{Speed}} \\
		&& Rec@50 & Rec@100 & Rec@50 & Rec@100 & \\
		\hline
		\multirow{5}{*}{VRD~\cite{visual_relationship}} 
		& LP~\cite{visual_relationship}  & 16.17 & 17.03 & 13.86 & 14.70 & 1.18$^*$\\
		& ViP-CNN~\cite{li2017vip} & 22.78 & 27.91 & 17.32 & 20.01 & 0.78\\
		& DR-Net~\cite{dai2017detecting} & 19.93 & 23.45 & 17.73 & 20.88 & 2.83 \\
		& ILC~\cite{plummer2016phrase} & 16.89 & 20.70 & 15.08 & 18.37 & 2.70$^{**}$\\
		& Ours Full:1-SMP & 25.90 & 30.52 & 18.16 & 21.04 & \textbf{0.45}\\
		& Ours Full:2-SMP & \textbf{26.03} & \textbf{30.77} & \textbf{18.32} & \textbf{21.20} & 0.55\\
		\hline
		\multirow{3}{*}{VG-MSDN~\cite{visual_genome,li2017scene}} 
		& ISGG~\cite{xu2017scene} & 15.87 & 19.45 & 8.23  & 	 10.88 & 1.64 \\
		& MSDN~\cite{li2017scene} & 19.95 & 24.93 & 10.72 & 14.22 & 3.56 \\
		& Ours-Full: 2-SMP & \textbf{22.84}  & \textbf{28.57} & \textbf{13.06} & \textbf{16.47} & \textbf{0.55} \\
		\hline
		\multirow{2}{*}{VG-DR-Net~\cite{visual_genome, dai2017detecting}} 
		& DR-Net~\cite{dai2017detecting} & 23.95 & 27.57 & \textbf{20.79} & 23.76 & 2.83\\
		& Ours-Full: 2-SMP & \textbf{26.91} & \textbf{32.63} & 19.88 & \textbf{23.95} & \textbf{0.55} \\
		\hline	
		\end{tabularx}
	\begin{tablenotes}\footnotesize
	$^*$ Only consider the post-processing time given the CNN features and object detection results. $^{**}$ As reported in \cite{plummer2016phrase}, it takes about 45 minutes to test 1000 images on single K80 GPU. 
	\end{tablenotes}
	\end{center}
	\label{tab:comparison}
\end{table*}

\subsection{Comparison with Existing Methods}
We compare our proposed F-Net with existing methods in Tab.~\ref{tab:comparison}. These methods can be roughly divided into two groups. One employs the two-stage pipeline, which is to detect the objects first and then recognize their relationships, including LP\cite{visual_relationship}, DR-Net~\cite{dai2017detecting} and ILC~\cite{plummer2016phrase}. Compared with these methods, our F-Net jointly recognizes the objects and their relationships, so the feature level connections can be leveraged for better recognition. In addition, complicated post-processing stages introduced by these methods may reduce the inference speed and make it more difficult to implement with GPU or other high-performance hardware like FPGA. 
The other methods like ViP-CNN~\cite{li2017vip}, ISGG~\cite{xu2017scene}, MSDN~\cite{li2017scene} adopt the similar pipeline to ours and propose different feature learning methods. Both ViP-CNN and ISGG used message passing to refine the object and predicate features. MSDN introduced an additional task, dense captioning, to improve scene graph generation. However, in these methods, each relationship is represented by an individual phrase feature. This leads to limited object proposals that are used to generate scene graph, as the number of relationships grows quadratically with that of the proposals. In comparison, our proposed subgraph-based pipeline significantly reduces the relationship representations by clustering them into subgraphs. 
Therefore, it allows us to use more object proposals to generate scene graph, and correspondingly, helps our model to perform better than these methods both in speed and accuracy. 

\section{Conclusion}

This paper introduces an efficient scene graph generation model, Factorizable Network~(F-Net). To tackle the problem of the quadratic combinations of possible relationships, a concise subgraph-based representation of the scene graph is introduced to reduce the number of intermediate representations during the inference. 2-D feature maps are used to maintain the spatial information within the subgraph region. Correspondingly, a Spatial-weighted Message Passing structure and a Spatial-sensitive Relation Inference module are designed to make use of the inherent spatial structure of the feature maps. Experiment results show that our model is significantly faster than the previous methods with better results. 

\section*{Acknowledgement}
This work is supported by Hong Kong Ph.D. Fellowship Scheme, SenseTime Group Limited, Samsung Telecommunication Research Institute, the General Research Fund sponsored by the Research Grants Council of Hong Kong (Project Nos. CUHK14213616, CUHK14206114, CUHK14205615, CUHK419412, CUHK14203015, CUHK14207814, CUHK14208417, CUHK14202217, and CUHK14239816), the Hong Kong Innovation and Technology Support Programme (No.ITS/121/15FX). 

\bibliographystyle{splncs04}
\bibliography{egbib}

\begin{thebibliography}{10}
\providecommand{\url}[1]{\texttt{#1}}
\providecommand{\urlprefix}{URL }
\providecommand{\doi}[1]{https://doi.org/#1}

\bibitem{antol2014zero}
Antol, S., Zitnick, C.L., Parikh, D.: Zero-shot learning via visual
  abstraction. In: ECCV (2014)

\bibitem{berg2012understanding}
Berg, A.C., Berg, T.L., Daume, H., Dodge, J., Goyal, A., Han, X., Mensch, A.,
  Mitchell, M., Sood, A., Stratos, K., et~al.: Understanding and predicting
  importance in images. In: CVPR (2012)

\bibitem{chang2014semantic}
Chang, A., Savva, M., Manning, C.: Semantic parsing for text to 3d scene
  generation. In: ACL (2014)

\bibitem{choi2010exploiting}
Choi, M.J., Lim, J.J., Torralba, A., Willsky, A.S.: Exploiting hierarchical
  context on a large database of object categories. In: CVPR (2010)

\bibitem{choi2013understanding}
Choi, W., Chao, Y.W., Pantofaru, C., Savarese, S.: Understanding indoor scenes
  using 3d geometric phrases. In: Proceedings of the IEEE Conference on
  Computer Vision and Pattern Recognition. pp. 33--40 (2013)

\bibitem{dai2017detecting}
Dai, B., Zhang, Y., Lin, D.: Detecting visual relationships with deep
  relational networks. CVPR  (2017)

\bibitem{das2013thousand}
Das, P., Xu, C., Doell, R.F., Corso, J.J.: A thousand frames in just a few
  words: Lingual description of videos through latent topics and sparse object
  stitching. In: CVPR (2013)

\bibitem{divvala2014learning}
Divvala, S.K., Farhadi, A., Guestrin, C.: Learning everything about anything:
  Webly-supervised visual concept learning. In: CVPR (2014)

\bibitem{elhoseiny2015sherlock}
Elhoseiny, M., Cohen, S., Chang, W., Price, B.L., Elgammal, A.M.: Sherlock:
  Scalable fact learning in images. In: AAAI (2017)

\bibitem{elliott2013image}
Elliott, D., Keller, F.: Image description using visual dependency
  representations. In: EMNLP (2013)

\bibitem{farhadi2010every}
Farhadi, A., Hejrati, M., Sadeghi, M.A., Young, P., Rashtchian, C.,
  Hockenmaier, J., Forsyth, D.: Every picture tells a story: Generating
  sentences from images. In: ECCV (2010)

\bibitem{fidler2007towards}
Fidler, S., Leonardis, A.: Towards scalable representations of object
  categories: Learning a hierarchy of parts. In: CVPR (2007)

\bibitem{galleguillos2010context}
Galleguillos, C., Belongie, S.: Context based object categorization: A critical
  survey. CVIU  (2010)

\bibitem{galleguillos2008object}
Galleguillos, C., Rabinovich, A., Belongie, S.: Object categorization using
  co-occurrence, location and appearance. In: CVPR (2008)

\bibitem{fast_rcnn}
Girshick, R.: Fast r-cnn. In: ICCV (2015)

\bibitem{gkioxari2015contextual}
Gkioxari, G., Girshick, R., Malik, J.: Contextual action recognition with r*
  cnn. In: ICCV (2015)

\bibitem{gong2014multi}
Gong, Y., Ke, Q., Isard, M., Lazebnik, S.: A multi-view embedding space for
  modeling internet images, tags, and their semantics. IJCV  (2014)

\bibitem{gould2008multi}
Gould, S., Rodgers, J., Cohen, D., Elidan, G., Koller, D.: Multi-class
  segmentation with relative location prior. IJCV  (2008)

\bibitem{guadarrama2013youtube2text}
Guadarrama, S., Krishnamoorthy, N., Malkarnenkar, G., Venugopalan, S., Mooney,
  R., Darrell, T., Saenko, K.: Youtube2text: Recognizing and describing
  arbitrary activities using semantic hierarchies and zero-shot recognition.
  In: ICCV (2013)

\bibitem{gupta2008beyond}
Gupta, A., Davis, L.S.: Beyond nouns: Exploiting prepositions and comparative
  adjectives for learning visual classifiers. In: ECCV (2008)

\bibitem{he2017mask}
He, K., Gkioxari, G., Doll{\'a}r, P., Girshick, R.: Mask r-cnn. In: ICCV (2017)

\bibitem{resnet}
He, K., Zhang, X., Ren, S., Sun, J.: Deep residual learning for image
  recognition. arXiv preprint arXiv:1512.03385  (2015)

\bibitem{hoiem2008putting}
Hoiem, D., Efros, A.A., Hebert, M.: Putting objects in perspective. IJCV
  (2008)

\bibitem{izadinia2014incorporating}
Izadinia, H., Sadeghi, F., Farhadi, A.: Incorporating scene context and object
  layout into appearance modeling. In: CVPR (2014)

\bibitem{densecap}
Johnson, J., Karpathy, A., Fei-Fei, L.: Densecap: Fully convolutional
  localization networks for dense captioning. arXiv preprint arXiv:1511.07571
  (2015)

\bibitem{johnson2015image}
Johnson, J., Krishna, R., Stark, M., Li, L.J., Shamma, D.A., Bernstein, M.S.,
  Fei-Fei, L.: Image retrieval using scene graphs. In: CVPR (2015)

\bibitem{karpathy2014deep}
Karpathy, A., Joulin, A., Fei-Fei, L.F.: Deep fragment embeddings for
  bidirectional image sentence mapping. In: NIPS (2014)

\bibitem{visual_genome}
Krishna, R., Zhu, Y., Groth, O., Johnson, J., Hata, K., Kravitz, J., Chen, S.,
  Kalantidis, Y., Li, L.J., Shamma, D.A., et~al.: Visual genome: Connecting
  language and vision using crowdsourced dense image annotations. IJCV  (2017)

\bibitem{imagenet_hinton}
Krizhevsky, A., Sutskever, I., Hinton, G.E.: Imagenet classification with deep
  convolutional neural networks. In: NIPS. pp. 1097--1105 (2012)

\bibitem{kulkarni2011baby}
Kulkarni, G., Premraj, V., Dhar, S., Li, S., Choi, Y., Berg, A.C., Berg, T.L.:
  Baby talk: Understanding and generating image descriptions. In: CVPR (2011)

\bibitem{kumar2010efficiently}
Kumar, M.P., Koller, D.: Efficiently selecting regions for scene understanding.
  In: CVPR (2010)

\bibitem{ladicky2010graph}
Ladicky, L., Russell, C., Kohli, P., Torr, P.H.: Graph cut based inference with
  co-occurrence statistics. In: ECCV (2010)

\bibitem{li2018visual}
Li, Y., Duan, N., Zhou, B., Chu, X., Ouyang, W., Wang, X., Zhou, M.: Visual
  question generation as dual task of visual question answering. In:
  Proceedings of the IEEE Conference on Computer Vision and Pattern
  Recognition. pp. 6116--6124 (2018)

\bibitem{li2017vip}
Li, Y., Ouyang, W., Wang, X., Tang, X.: Vip-cnn: Visual phrase guided
  convolutional neural network. CVPR  (2017)

\bibitem{li2017scene}
Li, Y., Ouyang, W., Zhou, B., Wang, K., Wang, X.: Scene graph generation from
  objects, phrases and region captions. In: ICCV (2017)

\bibitem{liao2017natural}
Liao, W., Shuai, L., Rosenhahn, B., Yang, M.Y.: Natural language guided visual
  relationship detection. arXiv preprint arXiv:1711.06032  (2017)

\bibitem{visual_relationship}
Lu, C., Krishna, R., Bernstein, M., Fei-Fei, L.: Visual relationship detection
  with language priors. In: ECCV (2016)

\bibitem{lu2018co-attending}
Lu, P., Li, H., Wei, Z., Wang, J., Wang, X.: Co-attending free-form regions and
  detections with multi-modal multiplicative feature embedding for visual
  question answering. In: AAAI (2018)

\bibitem{mensink2014costa}
Mensink, T., Gavves, E., Snoek, C.G.: Costa: Co-occurrence statistics for
  zero-shot classification. In: CVPR (2014)

\bibitem{relu}
Nair, V., Hinton, G.E.: Rectified linear units improve restricted boltzmann
  machines. In: ICML (2010)

\bibitem{peyre2017weakly}
Peyre, J., Laptev, I., Schmid, C., Sivic, J.: Weakly-supervised learning of
  visual relations. In: ICCV (2017)

\bibitem{plummer2016phrase}
Plummer, B.A., Mallya, A., Cervantes, C.M., Hockenmaier, J., Lazebnik, S.:
  Phrase localization and visual relationship detection with comprehensive
  linguistic cues. ICCV  (2017)

\bibitem{plummer2015flickr30k}
Plummer, B.A., Wang, L., Cervantes, C.M., Caicedo, J.C., Hockenmaier, J.,
  Lazebnik, S.: Flickr30k entities: Collecting region-to-phrase correspondences
  for richer image-to-sentence models. In: ICCV (2015)

\bibitem{rabinovich2007objects}
Rabinovich, A., Vedaldi, A., Galleguillos, C., Wiewiora, E., Belongie, S.:
  Objects in context. In: ICCV (2007)

\bibitem{visual_phrase_for_retrieval}
Ramanathan, V., Li, C., Deng, J., Han, W., Li, Z., Gu, K., Song, Y., Bengio,
  S., Rossenberg, C., Fei-Fei, L.: Learning semantic relationships for better
  action retrieval in images. In: CVPR (2015)

\bibitem{ramanathan2015learning}
Ramanathan, V., Li, C., Deng, J., Han, W., Li, Z., Gu, K., Song, Y., Bengio,
  S., Rossenberg, C., Fei-Fei, L.: Learning semantic relationships for better
  action retrieval in images. In: CVPR (2015)

\bibitem{regneri2013grounding}
Regneri, M., Rohrbach, M., Wetzel, D., Thater, S., Schiele, B., Pinkal, M.:
  Grounding action descriptions in videos. ACL  (2013)

\bibitem{faster_rcnn}
Ren, S., He, K., Girshick, R., Sun, J.: Faster r-cnn: Towards real-time object
  detection with region proposal networks. In: NIPS (2015)

\bibitem{rohrbach2015grounding}
Rohrbach, A., Rohrbach, M., Hu, R., Darrell, T., Schiele, B.: Grounding of
  textual phrases in images by reconstruction. arXiv preprint arXiv:1511.03745
  (2015)

\bibitem{rohrbach2013translating}
Rohrbach, M., Qiu, W., Titov, I., Thater, S., Pinkal, M., Schiele, B.:
  Translating video content to natural language descriptions. In: ICCV (2013)

\bibitem{russell2006using}
Russell, B.C., Freeman, W.T., Efros, A.A., Sivic, J., Zisserman, A.: Using
  multiple segmentations to discover objects and their extent in image
  collections. In: CVPR (2006)

\bibitem{sadeghi2011recognition}
Sadeghi, M.A., Farhadi, A.: Recognition using visual phrases. In: CVPR (2011)

\bibitem{salakhutdinov2011learning}
Salakhutdinov, R., Torralba, A., Tenenbaum, J.: Learning to share visual
  appearance for multiclass object detection. In: CVPR (2011)

\bibitem{VGG}
Simonyan, K., Zisserman, A.: Very deep convolutional networks for large-scale
  image recognition. arXiv preprint arXiv:1409.1556  (2014)

\bibitem{sivic2005discovering}
Sivic, J., Russell, B.C., Efros, A.A., Zisserman, A., Freeman, W.T.:
  Discovering objects and their location in images. In: ICCV (2005)

\bibitem{thomason2014integrating}
Thomason, J., Venugopalan, S., Guadarrama, S., Saenko, K., Mooney, R.:
  Integrating language and vision to generate natural language descriptions of
  videos in the wild. In: COLING (2014)

\bibitem{xiong2015recognize}
Xiong, Y., Zhu, K., Lin, D., Tang, X.: Recognize complex events from static
  images by fusing deep channels. In: CVPR (2015)

\bibitem{xu2017scene}
Xu, D., Zhu, Y., Choy, C.B., Fei-Fei, L.: Scene graph generation by iterative
  message passing. CVPR  (2017)

\bibitem{show_attend_tell}
Xu, K., Ba, J., Kiros, R., Cho, K., Courville, A., Salakhutdinov, R., Zemel,
  R.S., Bengio, Y.: Show, attend and tell: Neural image caption generation with
  visual attention. arXiv preprint arXiv:1502.03044  (2015)

\bibitem{yao2010grouplet}
Yao, B., Fei-Fei, L.: Grouplet: A structured image representation for
  recognizing human and object interactions. In: CVPR (2010)

\bibitem{yao2012describing}
Yao, J., Fidler, S., Urtasun, R.: Describing the scene as a whole: Joint object
  detection, scene classification and semantic segmentation. In: CVPR (2012)

\bibitem{yu2017visual}
Yu, R., Li, A., Morariu, V.I., Davis, L.S.: Visual relationship detection with
  internal and external linguistic knowledge distillation. ICCV  (2017)

\bibitem{zhang2017visual}
Zhang, H., Kyaw, Z., Chang, S.F., Chua, T.S.: Visual translation embedding
  network for visual relation detection. In: CVPR (2017)

\bibitem{zhuang2017towards}
Zhuang, B., Liu, L., Shen, C., Reid, I.: Towards context-aware interaction
  recognition. ICCV  (2017)

\bibitem{zitnick2013learning}
Zitnick, C.L., Parikh, D., Vanderwende, L.: Learning the visual interpretation
  of sentences. In: ICCV (2013)

\end{thebibliography}


\end{document}